\documentclass[letterpaper]{article} 
\usepackage{aaai25}  
\usepackage{times}  
\usepackage{helvet}  
\usepackage{courier}  
\usepackage[hyphens]{url}  
\usepackage{graphicx} 
\usepackage{natbib}  
\usepackage{caption} 
\usepackage[linesnumbered,ruled,vlined]{algorithm2e}
\usepackage{booktabs}
\usepackage{xcolor}
\usepackage{multirow}
\usepackage{todonotes}
\nocopyright
\usepackage{hyperref}
\title{LiveMind: Low-Latency Large Language Models with Simultaneous Inference}
\author{
    Chuangtao Chen\textsuperscript{\rm 1}, Grace Li Zhang\textsuperscript{\rm 2}, Xunzhao Yin\textsuperscript{\rm 3}, Cheng Zhuo\textsuperscript{\rm 3}, Ulf Schlichtmann\textsuperscript{\rm 1}, Bing Li\textsuperscript{\rm 4}
}
\affiliations{
    \textsuperscript{\rm 1}Technical University of Munich, Munich, Germany\\
    \textsuperscript{\rm 2}Technical University of Darmstadt, Darmstadt, Germany\\
    \textsuperscript{\rm 3}Zhejiang University, Hangzhou, China\\
    \textsuperscript{\rm 4}University of Siegen, Siegen, Germany

    \textsuperscript{\rm 1}\texttt{\{Chuangtao.chen, ulf.schlichtmann\}@tum.de}, \textsuperscript{\rm 2}\texttt{grace.zhang@tu-darmstadt.de} \\
    \textsuperscript{\rm 3}\texttt{\{xzyin1, czhuo\}@zju.edu.cn}, \textsuperscript{\rm 4}\texttt{bing.li@uni-siegen.de}
}

\usepackage{bibentry}

\begin{document}

\maketitle

\begin{abstract}
In this paper, we introduce LiveMind, a novel low-latency inference framework for large language model (LLM) inference which enables LLMs to perform inferences with incomplete user input. By reallocating computational processes to the input phase, a substantial reduction in latency is achieved, thereby significantly enhancing the interactive experience for users of LLMs. The framework adeptly manages the visibility of the streaming input to the model, allowing it to infer from incomplete user input or await additional content. Compared with traditional inference methods on complete user input, our approach demonstrates an average reduction in response latency of 84.0\% on the MMLU dataset and 71.6\% on the MMLU-Pro dataset, while maintaining comparable accuracy. Additionally, our framework facilitates collaborative inference and output across different models. By employing an large LLM for inference and a small LLM for output, we achieve an average 37\% reduction in response latency, alongside a 4.30\% improvement in accuracy on the MMLU-Pro dataset compared with the baseline. The proposed LiveMind framework advances the field of human-AI interaction by enabling more responsive and efficient communication between users and AI systems. Code and experimental results are available at \url{https://github.com/ChuangtaoChen-TUM/LiveMind}.
\end{abstract}\section{Introduction}
\label{sec:introduction}

Recently, large language models (LLMs) \cite{touvron2023llama, achiam2023gpt, bubeck2023sparks} based on the transformer architecture \cite{vaswani2017attention} have exhibited exceptional capabilities in natural language processing. These models, trained on extensive corpora and characterized by a substantial number of trainable parameters, have demonstrated powerful in-context learning abilities and exhibit significant potential across various domains \cite{boiko2023emergent,chen2021evaluating,liu2023chipnemo}. Nevertheless, the auto-regressive decoding nature and the considerable size of these LLMs result in slow response generation, thereby diminishing the quality of human-computer interaction. To enhance the inference speed of LLMs, various research efforts have been undertaken, including acceleration techniques such as pruning \cite{ma2023llm}, quantization \cite{lin2023awq,frantar2022gptq} and novel decoding methods \cite{chen2023accelerating,gloeckle2024better}.

\begin{figure*}[th]
  \centering
  \includegraphics[width=0.9\linewidth]{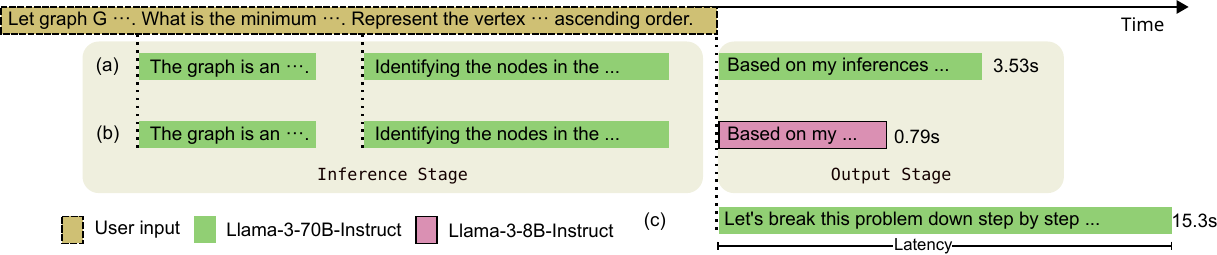}
  \caption{An example of the LiveMind framework. (a) LiveMind inference with Llama-3-70B-Instruct model; (b) LiveMind inference with Llama-3-70B-Instruct with Llama-3-8B-Instruct models; (c) Conventional inference on complete user input.}
  \label{fig:overview}
\end{figure*}

In this work, we identify another significant cause of high latency: the late processing of user input. Currently, most LLM interfaces, including text-based and audio-based LLMs \cite{rubenstein2023audiopalmlargelanguagemodel, borsos2023audiolmlanguagemodelingapproach}, only transmit the user's input to the model when the it is completed, which is usually indicated by a user-side event, such as pressing the enter key, or an interval of silence for audio case \cite{940814, shannon2017improved}. The LLM begins generating new tokens in an auto-regressive manner only after receiving the entire user input. During the period when the input is being provided, such as typing and speaking, the model remains in an idle status. Inspired by the concurrent inference observed in human conversations, where individuals process ongoing dialogues while listening, we propose that LLMs can similarly initiate inference with incomplete user input in an interactive scenario, where the input is streaming incrementally, \textit{e.g.}, typing or speaking. By leveraging the time during which the user is providing input, the model can pre-process these incomplete content. This approach can subsequently reduce the number of tokens required for inference once the complete input is received, thereby decreasing the latency perceived by users.

We introduce LiveMind, a framework that enables LLMs to process the user input concurrently with its streaming. Fig.~\ref{fig:overview} demonstrates an inference example with the proposed framework. As illustrated in Fig.~\ref{fig:overview}(a)(b), the majority of inferences can be conducted on incomplete user input during the input phase, which is also denoted as ``inference stage''. Utilizing these preliminary inferences, the final inference on the complete input to derive the final answer at the ``output stage'' can be significantly expedited. Since user-perceived latency is primarily determined by the duration of the final output stage, our framework substantially reduces this latency compared with the conventional inference based on the complete input in Fig.~\ref{fig:overview}(c). Additionally, by employing an powerful LLM during the inference stage and a small LLM during the output stage as shown in Fig.~\ref{fig:overview}(b), we can achieve an improvement in the overall accuracy without compromising the response latency. Accordingly, this work can lead to more dynamic and responsive AI systems, making interactions more seamless and efficient.

This paper is structured as follows: Sec.~\ref{sec:background} provides a more comprehensive overview of related work, including its benefits and limitations. In Sec.~\ref{sec:method}, we explain our proposed framework. The effectiveness of our method is evaluated by comparing it with the conventional inference method. The experimental results with an in-depth analysis are presented in Sec.~\ref{sec:experiment}. Finally, Sec.~\ref{sec:conclusion} concludes this paper.
\section{Background}
\label{sec:background}

\subsection{Related work}
\textbf{Simultaneous Translation with LLMs} A related area of research relevant to our work is the simultaneous translation with LLMs. Several studies have explored the application of LLMs to sequence-to-sequence (seq-to-seq) tasks that require simultaneous outputs, particularly in the context of machine translation \cite{wang2023simultaneous,agostinelli2023simul,koshkin2024transllama,wang2024conversational}. For instance, one approach involves prompting the LLM with a new request to translate each time when a set of new words arrive, and then selecting the output using a prefix selection algorithm \cite{wang2023simultaneous}. Another approach fine-tunes the model to output a special wait token, thereby achieving natural support for simultaneous output \cite{koshkin2024transllama}. In the task of simultaneous machine translation, the key of is to design a read-strategy which decides when to read partial input from the user and conduct an incremental processing and send the translation to the receiver.

In this study, we primarily concentrate on the scenario referred to as \textbf{simultaneous inference}, wherein LLMs are capable of processing streaming input within general interactive scenarios. It is similar to simultaneous seq-to-seq tasks, such as simultaneous machine translation and automatic speech recognition, processing starts when the user's input is not complete. 
However, simultaneous inference in general question answering tasks presents unique challenges compared to conventional seq-to-seq tasks. As there is no simple and direct mapping between the model's input and output as in seq-to-seq tasks, designing an optimal read strategy is challenging. Besides, general interactive scenarios such as question answering are more vulnerable to incomplete input due to the complex reasoning process required in these tasks. Consider that current LLMs are mostly fine-tuned dataset with complete user-LLM dialogues, feeding incomplete user input to the LLM can lead to degraded performance.

In this work, we propose LiveMind framework for simultaneous inference in interactive scenarios. This framework addresses the challenges associated with real-time interaction and processing in complex tasks that require multi-step reasoning. The framework enables the LLM to make intermediate inferences based on the user's partial input, and with these intermediate inferences, the number of reasoning steps required to obtain the final response when the complete user input is given can be reduced. Given that the computation time of LLMs is closely tied to the length of generated text \cite{zhong2024distserve, agarwal2023llm}, this framework significantly lowers user-perceived latency, particularly for complex tasks that necessitate multi-step reasoning.

We summarize our main contributions as follows:

\begin{itemize}
    \item We propose the LiveMind framework as a solution to reduce latency in general interactive scenarios. To the best of our knowledge, this represents the first attempt at implementing simultaneous inference of LLMs in general interactive scenarios.

    \item The framework features collaborative simultaneous inference, allowing the utilization of models in difference sizes in the inference and final output stages. For instance, employing an LLM with robust reasoning capabilities to make intermediate inference stages while a smaller and faster LLM to generate the final response. The collaborative simultaneous inference can enhance the task performance of the small LLM without compromising latency.

    \item We evaluated our proposed framework using various state-of-the-art open-source and closed-source LLMs, including Llama-3-8B-Instruct, Llama-3-70B-Instruct \cite{llama3modelcard}, and GPT-4o \cite{achiam2023gpt}, across diverse domain-specific questions. The results show significant improvements in latency with comparable accuracy. Specifically, our framework reduces latency by 84\% on the MMLU dataset \cite{mmlu_dataset, hendrycks2021ethics} and by 71.6\% on the MMLU-Pro dataset \cite{wang2024mmlu} using the Llama-3-70B model. For the GPT-4o Model, reductions of 45.0\% and 48.9\%  in latency can be achieved on the MMLU and the MMLU-Pro datasets respectively. Additionally, collaborative simultaneous inference with the Llama-3-70B-Instruct model as the inference model and Llama-3-8B-Instruct as the output model reduces latency by 37\% and increases accuracy by 4.30\% compared with Llama-3-8B baselines.
\end{itemize}
\section{Method}
\label{sec:method}
\begin{figure}[t]
    \centering
    \includegraphics[width=1.0\linewidth]{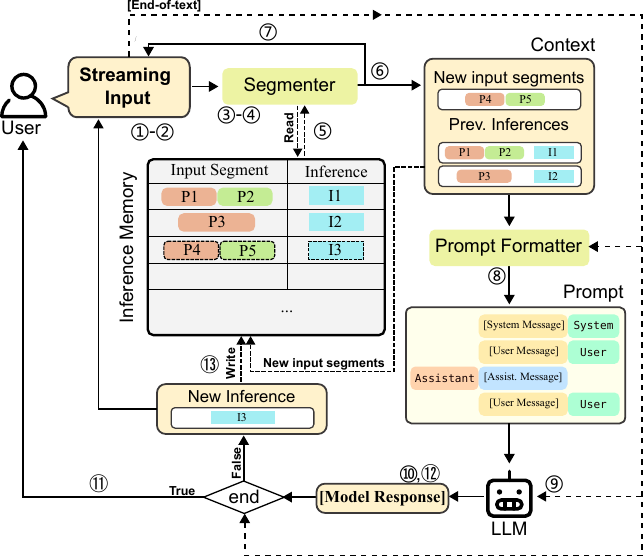}
    \caption{Architecture of the LiveMind framework, the circled numbers correspond to lines of Algorithm~\ref{alg1:livemind}.}
    \label{fig:live_mind_arch}
\end{figure}

In this section we explain the proposed LiveMind framework, including the key concept, the read strategy and the design of prompt formatter. Fig.~\ref{fig:live_mind_arch} and Algorithm~\ref{alg1:livemind} provides a synopsis of the proposed LiveMind framework. The details of the architecture and the algorithm will be explained in this section.

\newcommand\mycommfont[1]{\footnotesize\ttfamily\textcolor{blue}{#1}}
\SetCommentSty{mycommfont}
\SetKwInput{KwInput}{Input}                
\SetKwInput{KwOutput}{Output}              

\begin{algorithm}[th]
    \DontPrintSemicolon
    \KwInput{Streaming user input: $s$, End-of-text signal: $e$}
    \KwOutput{Response text to the user}
    \KwData{Inference model $\mathbf{L}_I$, Output model $\mathbf{L}_O$, Prompt formatter $\mathbf{P}$, \ Text segmenter $\mathbf{T}$, \ Inference Memory $\mathbf{M}$}
    
    \While{True}{
        s.\textit{update}()\;
        segments = $\mathbf{T}$(s)\;
        \tcc{drop the last unstable segment}
        segments = segments[0: -1]\;
        new\_segments, prev\_infer = $\mathbf{M}$.$read$(segments)\;
        \tcc{no new segments to process}
        \If{no new\_segments}{
            \textbf{continue}
        }
        prompt = $\mathbf{P}$(new\_segments, prev\_infer, $e$)\;

        \tcc{use the output model for final input, respond to the user}
        \If{$e$ is set}{
            response = $\mathbf{L}_O$(prompt)\;
            \Return response
        }

        \tcc{otherwise make a new inference and write it to memory}
        new\_inference = $\mathbf{L}_I$(prompt)\;
        $\mathbf{M}$.$write$(new\_segments, new\_inference)\;
    }
    
    \caption{Simultaneous inference process of the LiveMind framework}
    \label{alg1:livemind}
\end{algorithm}

\subsection{Key Concept}
LLMs have demonstrated the capability to tackle complicated tasks, especially with the use of the chain-of-thought (CoT) techniques \cite{wei2022chain,wang2022self}. The CoT method enhances the performance of LLMs to produce correct results, particularly in tasks necessitating multi-step reasoning, by enabling the decomposition of tasks into smaller, and sequential inferences. However, LLMs usually start the reasoning and deduce all intermediate results after the whole user input is available, although some intermediate results can already be obtained by partial input. This significantly delays the response of the LLMs in interaction-oriented systems and thus leads to a suboptimal user experience.

In this work, we propose to start the inference based on incomplete user input while the user is inputting, whether typing text or speaking audio. This preemptive inference obtains intermediate reasoning results in the background. These results are provided to the LLM together with the original complete user input as the user finishes the input. The LLM can then selectively skip reasoning steps and produce the final result quickly. Given that the majority of latency in LLMs arises from the auto-regressive decoding process \cite{zhong2024distserve, agarwal2023llm}, it is anticipated that this approach will mitigate the overall latency perceived by the user.

As illustrated in Fig.~\ref{fig:live_mind_arch}, the LiveMind framework comprises several key components that collaboratively facilitate simultaneous inference. The first component is the segmenter, which controls the reading actions within the framework. The inference memory serves to store segments of previous user inputs and inferences, maintaining context over time. The prompt formatter organizes this context retrieved from the memory, including user new input segments and previous inferences, into a structured prompt that facilitate the inference of the LLM. Finally, the LLM is responsible for generating new inferences and responding to the user.

\begin{figure}[t]
    \centering
    \includegraphics[width=0.9\linewidth]{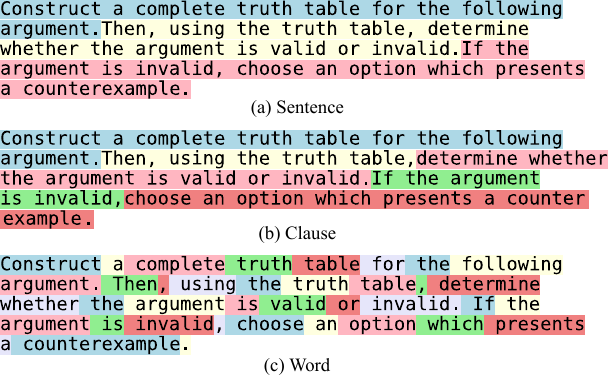}
    \caption{Text segmentation used by the segmenter in the LiveMind framework. An example text segmented by (a) sentence; (b) clause; (c) word.}
    \label{fig:segment}
\end{figure}

\subsection{Read Strategy}
Similar to other seq-to-seq tasks, we conceptualize the input as a continuous stream. In the context of text input, the data arrives incrementally, character by character. A critical issue arises regarding the optimal timing for transmitting the available data to the model. In the field of machine translation, the $wait$-$n$ strategy \cite{ma-etal-2019-stacl, agostinelli2023simul}, which involves accumulating a fixed number of words before processing them, is commonly used. However, unlike machine translation, where words and phrases often have direct counterparts in the target language, the significance of input words or tokens in general interactive tasks varies considerably in relation to the response. 

In the LiveMind framework, we address this issue by employing segmentation strategies based on syntactic structures rahter than fixed length. Specifically, we utilize a `sentence segmenter' and a `clause segmenter'. These methods capture more meaningful linguistic units compared to word-level segmentation, which operates at a finer granularity. The sentence segmenter is developed by modifying the sentence tokenizer from the Natural Language Toolkit (nltk) \cite{nltk}. For clause segmentation, sentences identified by the sentence segmenter are further divided using punctuation marks such as commas. To conduct ablation studies, we additionally include two finer segmenters that operate at the word and character levels. Fig.~\ref{fig:segment} demonstrates an example of text segmentation using segmenters with different granularities within the LiveMind framework. As the granularity increases from sentences to words, the frequency of reading actions by the language model rises, resulting in more fragmented content being processed each time.

In a streaming scenario where the whole input is constantly being updated character by character, the last segment from the partial input remains in flux. We assume that previous segments are stable and will not change, a condition that holds if the user only extends the input. We only send these stable segments for processing for consistency. To maintain consistency, only these stable segments are sent for processing. As demonstrated in Algorithm~\ref{alg1:livemind} lines 2-4, each time the input is updated, it is segmented. If multiple segments are available, we consider the earlier segments to be stable and unalterable by the user. On the contrary, the last segment, which may still be modified by additional user-input, is retained for subsequent processing.

\begin{figure}[t]
    \centering
    \includegraphics[width=\linewidth]{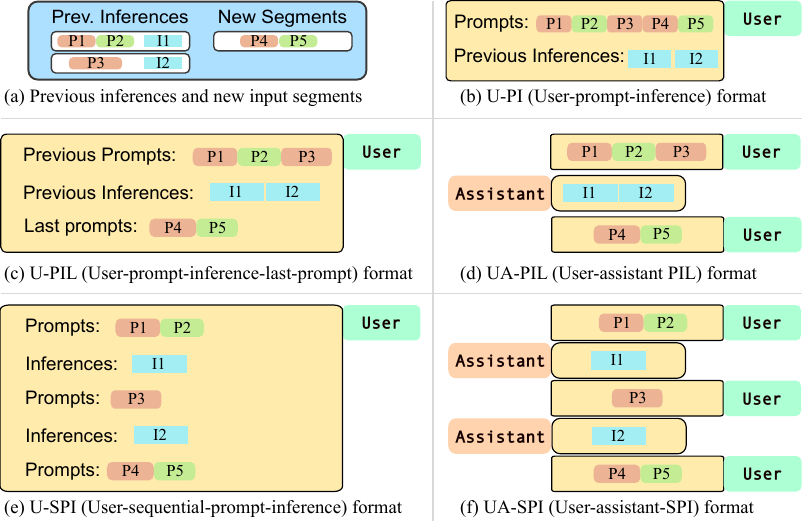}
    \caption{Five prompt-formats used by the LiveMind formatter: (a) previous inferences and new prompts; (b) U-PI format; (c) U-PIL format; (d) UA-PIL format; (e) U-SPI format; (f) UA-SPI format.}
    \label{fig:format}
\end{figure}

\subsection{Inference Memory}
In the LiveMind framework, intermediate results made by the LLM on incomplete user input are crucial fur subsequent inferences and responses. Therefore, a memory system to store the previous input segments and inferences is required. Besides, for computational efficiency, the framework should handle cases where users modify their previous inputs. 

The LiveMind framework features an inference memory as its center component to store the intermediate ``internal thoughts'' of the language model during simultaneous inference tasks. The memory comprises a list of saved entries, each contains one or multiple input segments and their corresponding inferences. When reading inferences from the memory, current input segments are matched with the saved input segments until the last entry or a conflict occurs. Conflicts can arise if the users delete or modify their previous input. In such cases, the memory's design functions as an inference cache that retrieves the inferences on the matched segments, therefore minimizing the cost for new inferences. For each read operation, the memory returns current \textbf{context}, which contains the matched entries and new input segments that do not match the stored entries, as shown in Fig.~\ref{fig:live_mind_arch}.

\subsection{Prompt Format}
Foundation models such as Llama \cite{touvron2023llama, llama3modelcard}, ChatGPT \cite{achiam2023gpt} are typically trained on extensive text corpus and then fine-tuned on dialog data to handle conversational tasks. These datasets consists of dialogues with complete inputs and responses. Therefore, applying these LLMs to the simultaneous inference tasks directly, where the user input is incomplete, can lead to confusion to the LLMs and thus result in degraded performance compared with the case in which the complete user input is given. To address this issue, a prompt formatter is developed within the LiveMind framework. This formatter organizes context content, including the new input segments and historical inferences retrieved from the inference memory (Algorithm~\ref{alg1:livemind} line~8), in a manner that LLM can comprehend without compromising task performance.

Most of LLMs that support conversation interactions require the following input format, where a list of texts assigned with a role label such as ``system'', ``user'' or ``assistant''. The text labelled with ``system'' is refereed to as the system message, which gives an overall control of the model's behavior. The ``user'' and ``assistant'' messages provide an history of user-LLM interactions. The list of texts with role labels are converted to a single string with special tokens before being processed by the LLM.

In the LiveMind framework, the prompt formatter have two main functions: (1) generating appropriate system message to inform the LLM with the task background and (2) organizing the retrieved context content into user and assistant messages in a structured format. For the system message, two options are provided for the inference stage and output stage, respectively. The current stage is controlled by the \texttt{[End-of-text]} signal as shown in Fig.~\ref{fig:live_mind_arch}. When the user has completed their input and is awaiting a response from the LLM, the signal is set, indicating the output stage. Conversely, when the user is still inputting, current stage is set as inference stage, where the model is prompted to make a temporary inference according to the new input segments or wait for additional input if necessary. 

For the context content, five distinct prompt formats are designed for the formatter: U-PI, U-SPI, UA-SPI, U-PIL and UA-PIL formats (\textbf{U}: User, \textbf{A}: Assistant, \textbf{P}: Prompt, \textbf{I}: Inference, \textbf{L}: Last prompts, \textbf{S}: Sequential). One of the five formats can be chosen to format the context information into the prompt. Fig.~\ref{fig:format} illustrates examples of these formats. Each message is labeled with its role, specifying whether it is a system, user or assistant message, to be sent to the LLM.

An example where two cached entries with two new input segments have been retrieved from the inference memory is given in Fig.~\ref{fig:format}~(a). Fig.~\ref{fig:format}(b)-(f) illustrate the different formats organize the context information. The U-PI format (Fig.~\ref{fig:format}~(b)) concatenates all available input segments and previous inferences into a single user message. In contrast, the U-PIL (Fig.~\ref{fig:format}~(c)) places the last unprocessed segments to the end of the message. The UA-PIL format is similar to U-PIL, with the distinction that the assistant, rather than the user, output previous inferences, as shown in Fig.~\ref{fig:format}~(d). The U-SPI format lists previous segments and inferences in a sequential and interleaving manner, with the unprocessed segments at the end. The UA-SPI format alternates between user messages for input segments and assistant messages and inferences, providing a step-by-step interaction history. Compared with the other formats, the SPI formats disrupt the continuity of previous input segments and inferences but offers the most detailed account of history structure.

\begin{table}[t]
\centering
\caption{Latency and accuracy of the LiveMind framework with the Llama-3-70B-Instruct as both the inference model and the output model on the MMLU dataset.}
\begin{tabular}{@{}cllcc@{}}
\toprule
              & \multicolumn{2}{c}{latency/s$\downarrow$ (speedup)}                 & \multicolumn{2}{c}{accuracy/\%$\uparrow$} \\ \midrule
p. format & \multicolumn{1}{c}{sentence} & \multicolumn{1}{c}{clause} & sentence                        & clause                  \\ \midrule
UA-PIL        & 0.88 (6.1$\times$)                  & 0.89 (6.0$\times$)                & 77.83                           & 77.93                   \\
UA-SPI        & \textbf{0.86 (6.3$\times$)}         & 0.86 (6.3$\times$)                & \textbf{78.03}                  & 77.54                   \\
U-PI          & 0.89 (6.0$\times$)                  & 0.94 (5.7$\times$)                & 78.03                           & 77.54                   \\
U-PLI         & 1.25 (4.3$\times$)                  & 1.34 (4.0$\times$)                & 76.95                           & 75.68                   \\
U-SPI         & \textbf{1.18 (4.6$\times$)}         & 1.18 (4.6$\times$)                & \textbf{78.32}                  & 75.78                   \\ \midrule
Baseline      & \multicolumn{2}{c}{5.39}                                  & \multicolumn{2}{c}{79.39}                                 \\ \bottomrule
\end{tabular}
\label{tab:70b_mmlu}
\end{table}

\subsection{LLM Inference}
Once the information is formatted into a comprehensible prompt for the LLM, it is sent to the model for processing. As detailed in Algorithm~\ref{alg1:livemind}, lines 10 and 12, an inference model $\textbf{L}_I$ and an output model $\textbf{L}_O$ are employed during the inference stage and the output stage, respectively. When this \texttt{[End-of-text]} signal $e$ is set, indicating that the user has completed input, the output model $\textbf{L}_O$ generates the final response and sends it back to the user (lines 10-11) and the algorithm finishes. Otherwise, a new inference is made by the inference model $\textbf{L}_I$ on the new input segments, and both the new segments and the inference are updated in the memory (lines 12-13).

The inference model and the output model can either be the same LLM or different LLMs. When they collaborates, the major cause of user-perceived latency comes from the generation process of the output model. Therefore, a larger, powerful LLM can be used as the inference model, while a smaller, faster LLM can be employed to generate the final response. This approach allows the smaller model to leverage the superior capabilities of the larger model's previous inferences, thereby enhancing the performance of the small LLM without compromising the latency.
\section{Experimental Results}
\label{sec:experiment}
This section presents a series of experiments designed to evaluate the efficacy of the proposed LiveMind framework. In Sec.~\ref{subsec:experiment_setup}, the experimental settings, including the datasets, models, and metrics employed are detailed. The results of the LiveMind framework on accuracy and latency in interactive scenarios on the MMLU dataset \cite{mmlu_dataset, hendrycks2021ethics} and the MMLU-Pro \cite{wang2024mmlu} datasets are presented in Sec.~\ref{subsec:real_time_eval}. An analysis of the impact of different segmentation methods are given in Sec.~\ref{subsec:seg}. Finally, we evaluated the computational cost associated with the LiveMind framework and the efficacy of the model collaboration feature in Sec.~\ref{subsec:comp_cost} and Sec.~\ref{subsec:model_colab} respectively.

\begin{table}[t]
\centering
\caption{Latency and accuracy of the LiveMind framework with the Llama-3-70B-Instruct as both inference model and output model model on the MMLU-Pro dataset.}
\begin{tabular}{@{}ccccc@{}}
\toprule
& \multicolumn{2}{c}{latency(s)$\downarrow$ (speedup$\times$)} & \multicolumn{2}{c}{accuracy(\%)$\uparrow$} \\ \midrule
p. format & sentence           & clause             & sentence        & clause       \\ \midrule
UA-PIL        & \textbf{3.38} (3.1$\times$)        & 3.26 (3.2$\times$)        & \textbf{57.62}           & 56.93        \\
UA-SPI        & 3.08 (3.4$\times$)        & \textbf{2.96} (3.5$\times$)        & 56.84           & \textbf{57.13}        \\
U-PI          & 4.97 (2.1$\times$)        & 5.00 (2.1$\times$)        & 55.08           & 55.47        \\
U-PLI         & 4.94 (2.1$\times$)        & 4.92 (2.1$\times$)        & 56.54           & 56.74        \\
U-SPI         & 4.91 (2.1$\times$)        & 4.69 (2.2$\times$)        & 56.35           & 55.86        \\ \midrule
Baseline      & \multicolumn{2}{c}{10.45}               & \multicolumn{2}{c}{57.42}      \\ \bottomrule
\end{tabular}
\label{tab:70b_mmlu_pro}
\end{table}

\subsection{Experiment setup}
\label{subsec:experiment_setup}
\textbf{Dataset \& Models} We selected the MMLU dataset \cite{mmlu_dataset, hendrycks2021ethics} and the MMLU-Pro dataset \cite{wang2024mmlu} to evaluate our proposed framework because they contain questions from various domains and of a range of difficulty levels. The MMLU (Measuring Massive Multitask Language Understanding) dataset consists of multiple-choice questions across 57 categories, each with four choices. The MMLU-Pro dataset is more robust and challenging compared with the standard MMLU dataset \cite{wang2024mmlu}, containing 12,032 questions across 14 categories, each formatted as a multiple-choice question with 3 to 10 possible answers. The MMLU-Pro dataset features more choices, longer question contexts, and higher difficulty. For all evaluations, we employed a zero-shot approach and reported the pass@1 results. For the experiments, we randomly selected 1024 questions from the corresponding dataset. 

The framework was evaluated using various SOTA LLMs, encompassing both open-source and closed-source variants. For the open-sourced models, we employed the Llama-3-70B-Instruct and Llama-3-8B-Instruct models. These models were executed with 4-bit quantization using AutoAWQ \cite{lin2023awq} on the vLLM platform \cite{kwon2023efficient} to expedite the inference. All local experiments were conducted on a single Nvidia A100-80G GPU, with the temperature parameter set to 0 to ensure reproducible results. 

\textbf{Streaming input simulation}: To simulate human interaction with LLMs, we developed a text stream that delivers characters of the questions from the datasets sequentially at a configurable speed. The input speed of the stream was set to mimic the average human typing speed, which is approximately 240 characters per minute. The latency for each question was measured as the time difference between the availability of the entire user input and the generation of the complete response.

\subsection{Real-time latency evaluation}
\label{subsec:real_time_eval}
Table~\ref{tab:70b_mmlu} presents the evaluation results of the proposed LiveMind framework using the Llama-3-70B-Instruct model on the MMLU dataset. The LiveMind framework significantly reduces latency through simultaneous inference compared to the baseline, which averages 5.39 seconds per question. For instance, using the UA-SPI prompt and sentence level granularity has an average latency of 0.86 seconds, which is a $6.3\times$ speedup, with only a slight 1.36\% decrease in accuracy. On the MMLU-Pro dataset as shown in Table~\ref{tab:70b_mmlu_pro}, which comprises more difficult tasks than the MMLU dataset, the framework achieves up to a 3.5$\times$ speedup with comparable or improved accuracy. Specifically, the UP-PIL prompt format with sentence level segmentation results in a 3.1$\times$ speedup with an increase in accuracy compared with the baseline inference.

The LiveMind framework was also evaluated using the SOTA closed-source model GPT-4o \cite{achiam2023gpt}. The results for the MMLU and MMLU-Pro datasets are presented in Table~\ref{tab:gpt4o_mmlu} and Table~\ref{tab:gpt4o_mmlu_pro}. On both datasets, our method can reduce the latency by up to $50\%$ while achieving similar or higher accuracy.

\subsection{Segmentation Granularity}
\label{subsec:seg}
As shown in Table~\ref{tab:70b_mmlu} and Table~\ref{tab:70b_mmlu_pro}, both the sentence level and clause-level segmentations result in significant latency reduction and comparable accuracy compared with the baseline. In this section, we will provide an in-depth analysis of how segmentation granularity impacts model performance and latency, including additional evaluations with word-level and character-level segmentations.

\begin{figure}[t]
    \centering
    \includegraphics[width=0.85\linewidth]{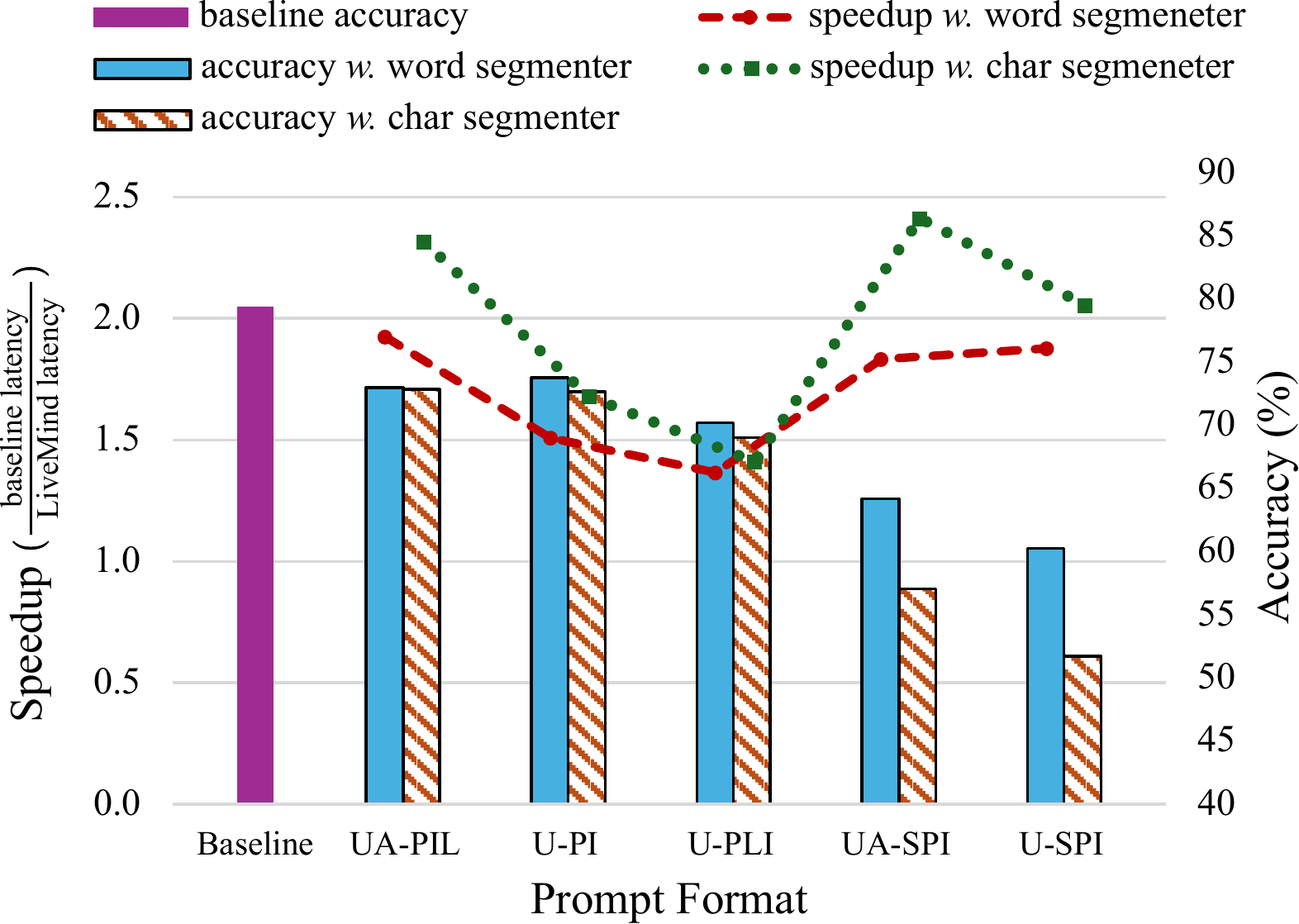}
    \caption{Latency (speedup) and accuracy of the LiveMind framework using word and character segmenters with Llama-3-70B-Instruct as both the inference model and the output model on the MMLU dataset.}
    \label{fig:70b_mmlu_pro_word_char}
\end{figure}

Fig.~\ref{fig:70b_mmlu_pro_word_char} presents the experimental results of Llama-3-70B-Instruct model on MMLU dataset using smaller segmentations within the LiveMind framework. While these segmentations still demonstrate reduction in terms of latency when compared with the baseline, they significantly lower accuracy, particularly in the UA-SPI and U-SPI formats. As detailed in Sec.~\ref{sec:method}, SPI (Sequential-Prompt-Inference) involves presenting historical prompts and inferences to the LLM in an interleaving manner. As current SOTA LLMs are trained on tasks where the prompts are given complete, these formats can fragment the interaction history, leading to confusion. Conversely, the PIL and PLI formats concatenate all previous prompts, presenting the history in a continuous manner, which facilitates easier comprehension. 

\begin{table}[t]
\centering
\caption{Latency and accuracy of the LiveMind framework with the GPT-4o as both the inference model and the output model on the MMLU dataset.}
\begin{tabular}{@{}ccccc@{}}
\hline
 & \multicolumn{2}{c}{latency(s)$\downarrow$ (speedup$\times$)} & \multicolumn{2}{c}{accuracy(\%)$\uparrow$} \\ \hline
p. format & sentence                    & clause             & sentence       & clause         \\ \hline
UA-PIL    & 1.99 (1.8$\times$)          & 1.97 (1.8$\times$) & 87.70 & 88.28          \\
UA-SPI    & \textbf{1.98} (1.8$\times$) & 2.01 (1.8$\times$) & \textbf{87.79}          & 88.18 \\ \hline
Baseline  & \multicolumn{2}{c}{3.60}                         & \multicolumn{2}{c}{88.48}       \\ \hline
\end{tabular}
\label{tab:gpt4o_mmlu}
\end{table}

\begin{table}[t]
\centering
\caption{Latency and accuracy of the LiveMind framework with the GPT-4o as both the inference model and the output model on the MMLU-Pro dataset.}
\begin{tabular}{@{}ccccc@{}}
\hline
 & \multicolumn{2}{c}{latency(s)$\downarrow$ (speedup$\times$)} & \multicolumn{2}{c}{accuracy(\%)$\uparrow$} \\ \hline
p. format & sentence                    & clause             & sentence       & clause         \\ \hline
UA-PIL    & 4.33 (1.7$\times$)          & 3.69 (2.0$\times$) & 74.80 & 73.83          \\
UA-SPI    & \textbf{3.80} (2.0$\times$) & 3.86 (1.9$\times$) & \textbf{74.90}          & 73.14 \\ \hline
Baseline  & \multicolumn{2}{c}{7.43}                         & \multicolumn{2}{c}{73.93}       \\ \hline
\end{tabular}
\label{tab:gpt4o_mmlu_pro}
\end{table}

\subsection{Computational Cost}
\label{subsec:comp_cost}
In this section, an analysis of the computational cost with the proposed LiveMind framework is presented. Due to the inherent characteristics of the transformer architecture, computational speed varies across different processing stages. For example, while prefill processing usually involves a large amount of FLOPs, it achieves a higher token processing speed compared with auto-regressive decoding, since the prefill tokens can be processed in parallel. Consequently, in our experiments, we measured the total computation time - defined as the duration for which the GPU is occupied - as the primary metric for computational cost.

\begin{figure}[t]
    \centering
    \includegraphics[width=\linewidth]{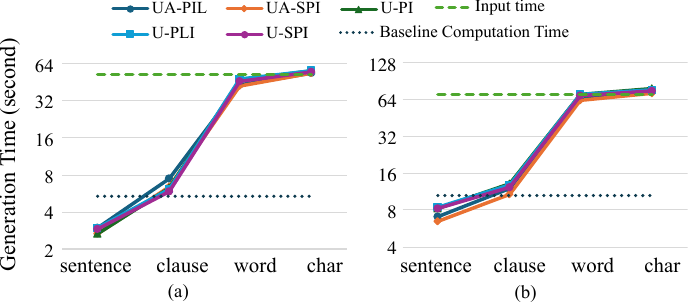}
    \caption{Average generation time of Llama-3-70B-Instruct model using the LiveMind framework with different settings on (a) MMLU dataset; (b) MMLU-Pro dataset.}
    \label{fig:gen_time_seg}
\end{figure}

As discussed in Sec.~\ref{sec:method}, the LiveMind framework enables the model to commence inference while the user is generating input. This results in the total computation time being longer than the waiting time perceived by the user. Fig.~\ref{fig:gen_time_seg} illustrates the average generation time of running Llama-3-70B-Instruct under various LiveMind framework settings on the MMLU and MMLU-Pro datasets. The dotted line represent the computation time of the baseline, which is identical to the baseline latency. The dashed line represents the user-input time, reflecting the duration the user spends typing the question.

Figure~\ref{fig:gen_time_seg} demonstrates that segmentation granularity has a more significant impact on the generation time than the prompt format. For example, sentence-level segmentation results in the lowest computational cost, even lower than the baseline. This is because, despite the increased number of response steps introduced by the simultaneous inference framework, the model performs incremental inference on new input segments without requiring the chain-of-thought technique. In contrast, finer segmentations, such as word and character levels, substantially increase computational cost, approaching the total input time. This occurs because the model is frequently invoked by new words and characters from the user, with little idle time waiting for more content. Improving the performance and computational efficiency simultaneous inference of LLMs on these fine granularities remains a challenging task.

\subsection{Model Collaboration}
\label{subsec:model_colab}
As discussed in Sec.~\ref{sec:method}, the LiveMind framework consists of two stages: the inference stage, where the LLM makes temporary inferences based on available user input, and the output stage, where the LLM responds to the user with inferences it has made. This design enable the utilization of different LLMs at each stage to leverage their unique strengths, such as the strong reasoning capabilities of the larger LLMs and quick response time of the smaller LLMs.

We conducted experiments to study the collaboration between different models. For baseline setting, we used the Llama-3-8B-Instruct model on complete user input. In the LiveMind framework, we employed the Llama-3-8B-Instruct as the output model to minimize the response latency, while a more powerful Llama-3-70B-Instruct model was used as the inference model to generate intermediate inferences. It is anticipated that the performance of the small LLM can be strengthened by the large LLM without the detriment of response latency.

Table~\ref{tab:model_colab} presents the experimental results of model collaboration between the Llama-3-70B-Instruct and Llama-3-8B-Instruct models on the MMLU-Pro dataset. Compared with the baseline, which uses the Llama-3-8B-Instruct model on the complete input, the collaborative approach demonstrates improved accuracy across all configurations. Specifically, the sentence-level segmentation with UA-SPI prompt format achieves an average latency of 0.83 seconds with an accuracy of 41.31\%, showing significant enhancements in both response speed and accuracy. For simultaneous inference with Llama-3-8B-Instruct as both the inference model and the output model, we list the best results (UA-SPI (8B)) in Table~\ref{tab:model_colab}. Compared with the inference that solely uses Llama-3-8B-Instruct, the collaborative inference also demonstrates improved task performance without compromising latency. The experimental results indicate that the smaller Llama-3-8B-Instruct model benefit from the capability of the Llama-3-70B-Instruct model by utilizing the its inferences.

\begin{table}[t]
\centering
\caption{Model collaboration of Llama-3-70B-Instruct as the inference model and Llama-3-8B-Instruct as the output model on the MMLU-Pro dataset.}
\begin{tabular}{ccccc}
\hline
                       & \multicolumn{2}{c}{latency(s)} & \multicolumn{2}{c}{accuracy(\%)} \\ \hline
\multicolumn{1}{c}{prompt format} & sentence       & clause       & sentence        & clause       \\ \hline
UA-PIL                            & 0.99           & 1.22         & 39.75           & 39.55        \\
UA-SPI                            & \textbf{0.83}           & 0.83         & \textbf{41.31}           & 39.06        \\
U-PI                              & 1.24           & 1.13         & 40.04           & 39.45        \\
U-PLI                             & 1.19           & 1.35         & 39.65           & 38.18        \\
U-SPI                             & 1.08           & 1.32         & 41.31           & 37.60        \\ \hline
UA-SPI (8B)        & 0.89     & 0.65 & 37.99        & 36.91           \\ \hline
Baseline                          & \multicolumn{2}{c}{1.32}      & \multicolumn{2}{c}{37.01}      \\ \hline
\end{tabular}
\label{tab:model_colab}
\end{table}
\section{Conclusion}
\label{sec:conclusion}

In this paper, we introduce LiveMind, a framework designed to enable low-latency LLMs with simultaneous inference capabilities. To the best of our knowledge, this represents the first attempt to implement simultaneous inference in LLMs for general interactive scenarios. Our experimental results validate the effectiveness of the proposed simultaneous inference method, suggesting that this is a promising direction for further research.

\bibliography{references}

\end{document}